\pdfoutput=1

\documentclass{article}

\usepackage{arxiv}

\usepackage[utf8]{inputenc} 
\usepackage{hyperref}       
\usepackage{url}            
\usepackage{booktabs}       
\usepackage{amsfonts}       
\usepackage{nicefrac}       
\usepackage{microtype}      
\usepackage[title]{appendix}
\usepackage{graphicx}

\usepackage{array}
\usepackage{multirow}
\usepackage{amsmath}

\usepackage{doi}
\usepackage{etoolbox}
\usepackage[natbibapa]{apacite}
\AtBeginDocument{}
\renewcommand{\APACrefnote}[1]{}
\newtoggle{bibdoi}
\newtoggle{biburl}
\makeatletter
\newsavebox{\bib@url}
\newsavebox{\bib@doi}

\undef{\APACrefURL}
\undef{\endAPACrefURL}
\undef{\APACrefDOI}
\undef{\endAPACrefDOI}

\newenvironment{APACrefURL}
  {\global\toggletrue{biburl}\lrbox\bib@url}
  {\endlrbox}

\newenvironment{APACrefDOI}
  {\global\toggletrue{bibdoi}\lrbox\bib@doi}
  {\endlrbox}

\newcommand{\printinfo}{
  \iftoggle{bibdoi}{\usebox{\bib@doi}}{\usebox{\bib@url}}
  \togglefalse{bibdoi}
}

\AtBeginEnvironment{thebibliography}{
\pretocmd{\PrintBackRefs}{%
  \iftoggle{bibdoi}
    {\iftoggle{biburl}{\unskip\unskip}{}\usebox{\bib@doi}}
    {\iftoggle{biburl}{Retrieved from \usebox{\bib@url}}}{}
  \togglefalse{bibdoi}\togglefalse{biburl}%
}{}{}}
\makeatother

\title{Flood-DamageSense: Multimodal Mamba with Multitask Learning for Building Flood Damage Assessment using SAR Remote Sensing Imagery}

\date{} 					

\renewcommand{\headeright}{}
\renewcommand{\undertitle}{}
\renewcommand{\shorttitle}{}

\hypersetup{
pdftitle={},
pdfsubject={},
}

\begin{document}
\maketitle

\begin{center}
{\Large
Yu-Hsuan Ho\textsuperscript{a,*},
Ali Mostafavi\textsuperscript{a}
\par}

\bigskip
\textsuperscript{a} Urban Resilience.AI Lab, Zachry Department of Civil and Environmental Engineering,\\ Texas A\&M University, College Station, TX\\
\vspace{6pt}
\textsuperscript{*} corresponding author, email: yuhsuanho@tamu.edu
\\
\end{center}
\bigskip
\begin{abstract}
Most post-disaster damage classifiers succeed only when destructive forces leave clear spectral or structural signatures—conditions rarely present after inundation. Consequently, existing models perform poorly at identifying flood-related building damages. The model presented in this study, Flood-DamageSense, addresses this gap as the first deep-learning framework purpose-built for building-level flood-damage assessment. The architecture fuses pre- and post-event SAR/InSAR scenes with very-high-resolution optical basemaps and an inherent flood-risk layer that encodes long-term exposure probabilities, guiding the network toward plausibly affected structures even when compositional change is minimal. A multimodal Mamba backbone with a semi-Siamese encoder and task-specific decoders jointly predicts (1) graded building-damage states, (2) floodwater extent, and (3) building footprints. Training and evaluation on Hurricane Harvey (2017) imagery from Harris County, Texas—supported by insurance-derived property-damage extents—show a mean F1 improvement of up to 19 percentage points over state-of-the-art baselines, with the largest gains in the frequently misclassified “minor” and “moderate” damage categories. Ablation studies identify the inherent-risk feature as the single most significant contributor to this performance boost. An end-to-end post-processing pipeline converts pixel-level outputs to actionable, building-scale damage maps within minutes of image acquisition. By combining risk-aware modeling with SAR’s all-weather capability, Flood-DamageSense delivers faster, finer-grained, and more reliable flood-damage intelligence to support post-disaster decision-making and resource allocation.
\end{abstract}

\keywords{ Rapid damage assessment \and Mamba \and Synthetic Aperture Radar (SAR) \and Multimodal learning \and Multitask learning \and Remote sensing \and Artificial intelligence \and Disaster response}


\section{Introduction}
\label{sec:introduction}
Rapid, building-level assessment of flood damage is pivotal for activating emergency response, directing resources, and hastening recovery, yet traditional field inspections are slow, costly, and can place personnel in harm’s way \citep{SPENCER2019199}. Motivated by these constraints, researchers have increasingly adopted automated techniques for flood risk analysis \citep{ho2024elev,ho2025integrated,huang2025high}, prediction \citep{dong2021hybrid,esparza2025improving}, and loss assessment \citep{dong2020bayesian,nofal2024community}. Machine-learning models excel at extracting complex, non-linear patterns from large data sets \citep{rafiei2017novel,huang2025high}, whereas Bayesian frameworks provide principled uncertainty quantification \citep{dong2020bayesian,wang2024scalable,perez2019recurrent}. Despite these advances, research that delivers near-real-time flood damage intelligence at the individual-structure scale remains nascent, largely because observational data are scarce during and immediately after inundation events.

Satellite imagery offers the dual advantages of rapid availability and continent-scale coverage. The release of the xBD dataset \citep{Gupta_2019_CVPR_Workshops}, which contains pre- and post-disaster satellite imagery with building polygons and ordinal damage labels from various catastrophic events, has spurred the development of promising multi-hazard damage-mapping pipelines using optical sensors \citep{kaur2023large,braik2024automated}. However, clouds, commonly present during floods, obscure targets at precisely the moment information is most needed. Similar visibility problems in other fields—e.g., concealed concrete defects detected by infrared thermography \citep{sirca2018infrared} or subtle brain anomalies revealed by MRI \citep{mirzaei2018segmentation,nogay2020machine}—have been overcome by deploying sensing modalities better suited to the task. In Earth observation, Synthetic Aperture Radar (SAR) offers an all-weather, day-night alternative, and SAR intensity and InSAR (Interferometric Synthetic Aperture Radar) coherence are well established for floodwater mapping \citep{singh2021review,amitrano2024flood}. Yet coarse native resolution and speckle noise have limited SAR’s adoption for building-level flood-damage assessment \citep{cao2020building}. Hybrid approaches that fuse SAR with optical data \citep{rudner2019multi3net} or embed SAR in Bayesian causal networks \citep{wang2024scalable} show promise but either inherit the weather constraints of optical imagery or rely on resource-intensive inference.

More broadly, satellite change-detection research has progressed from convolutional Siamese U-Nets \citep{SiameseUNet} through Transformer architectures \citep{dosovitskiy2020image,vaswani2017attention,chen2022bit,bandara2022changeformer,kaur2023large} to the recent family of Mamba-state space models, whose linear complexity affords efficient long-range context modeling \citep{gu2023mamba,liu2024vmamba,zhu2024vim}. Remote-sensing variants, such as ChangeMamba \citep{chen2024changemamba}, have begun to exploit these characteristics for general change detection and multi-hazard damage mapping, yet none target the subtle, low-contrast signatures typical of flood impacts. Specialized, hazard-specific models underscore the value of tailored feature design: wind and earthquake studies combine aerial, street-view or SAR data to capture phenomena unique to those events \citep{cheng2021deep,khajwal2023post,gu2025multi,wu2025earthquake,xiao2025damagecat,xiao2025perspective}. For floods, however, existing satellite pipelines rarely progress beyond binary damage labels and often depend on hand-annotated ground truth that is both labor-intensive and error-prone \citep{rudner2019multi3net,cao2020building}. Moreover, most vision papers stop at pixel-level outputs; only a few workflows attempt to translate model predictions into georeferenced, building-level maps, and those that do typically extract isolated image chips per footprint, thereby discarding contextual cues and introducing additional computation \citep{braik2024automated}. Public building-footprint databases \citep{USBuildingFootprints,GlobalMLBuildingFootprints,OpenBuildings} remain an untapped asset for tying raster predictions directly to individual structures and for streamlining the image-to-map pipeline.

These gaps motivate Flood-DamageSense, the first deep-learning framework purpose-built for multi-level, building-specific flood-damage assessment. The system employs a multimodal Mamba backbone that ingests paired pre-/post-event SAR–InSAR scenes, pre-event very-high-resolution (VHR) optical basemaps, and a historical flood-risk surface derived from topography, hydrology, and insurance-claim records \citep{mobley2021quantification}. Floodwater segmentation is introduced as an auxiliary task to regularize learning and impose hydrological consistency, while supervision is provided by property-damage-extent (PDE) labels compiled from National Flood Insurance Program (NFIP) and Federal Emergency Management Agency (FEMA) Individual Assistance (IA) payouts \citep{LIU2024FloodDamageCast}. Coupled with a building-footprint overlay that converts pixel probabilities to per-structure scores, the pipeline yields georeferenced damage maps within minutes of SAR acquisition, operates under full cloud cover, preserves fine-scale semantic information, and eliminates the need for manual labeling. A case study on Hurricane Harvey (Harris County, Texas, 2017) demonstrated that Flood-DamageSense accurately discriminated minor and moderate losses—categories where prior optical-centric models most often fail—thereby filling a critical void in the disaster-response toolbox.

\section{Methodology}
\label{sec:methodology}
\begin{figure}[ht]
    \centering
    \includegraphics[width=\textwidth]{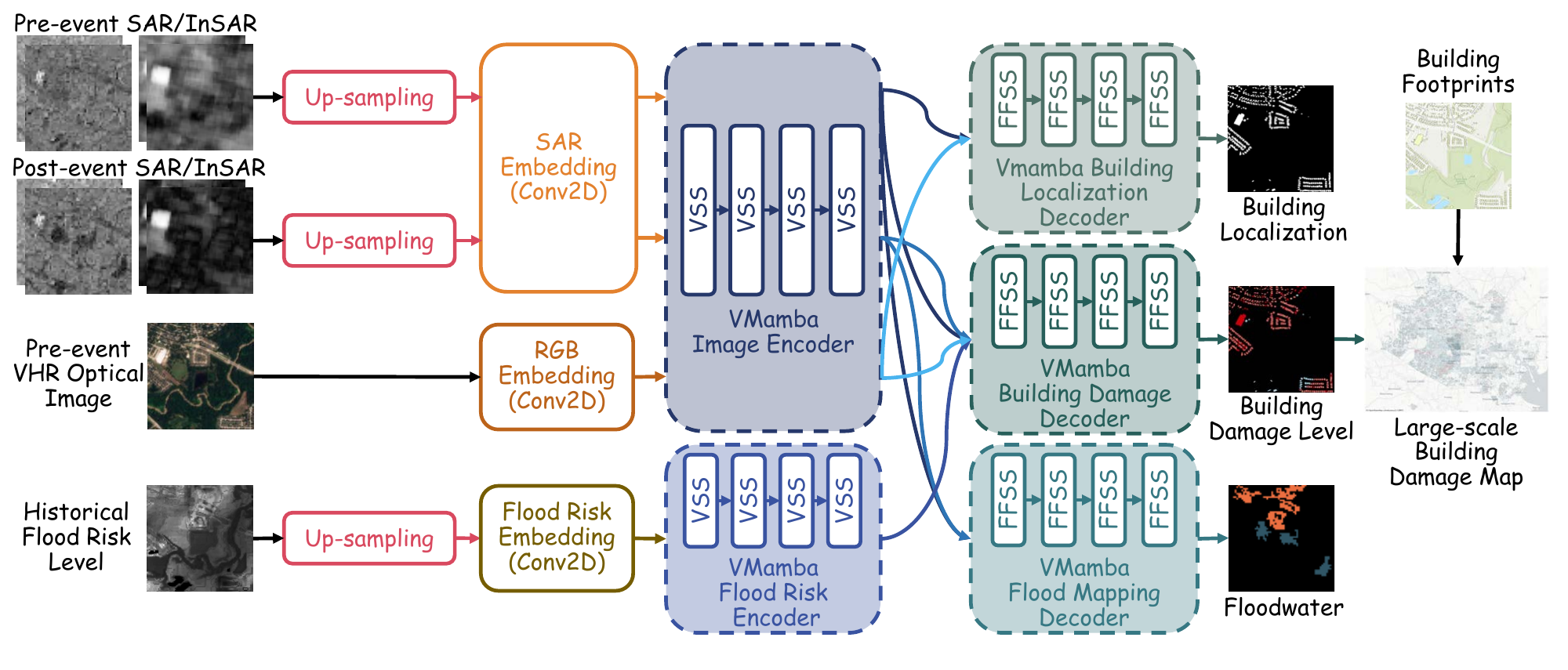}
    \caption{Architecture of Flood-DamageSense. The framework processes multimodal inputs, including pre- and post-event SAR and InSAR imagery, pre-event VHR optical imagery, and historical flood risk levels through initial embedding layers followed by VMamba encoders. A multitask learning approach utilizes distinct decoders to concurrently generate building localization masks, building damage levels, and floodwater maps. The pixel-level building damage outputs are subsequently integrated with georeferenced building footprints to produce the final large-scale building damage map.}
    \label{fig:conceptual_figure}
\end{figure}

This section elaborates on the proposed Flood-DamageSense framework, designed for rapid building-level flood damage assessment by primarily utilizing SAR remote sensing imagery to overcome the weather-related limitations inherent in optical sensors. The framework employs a multimodal Mamba-based architecture that leverages multitask learning. Multimodal inputs include SAR imagery to maintain robustness to weather conditions, historical flood risk levels to enhance perception to damage not involving structural changes, and pre-event VHR optical satellite imagery to refine spatial details. To further bolster the performance of the building damage assessment, floodwater mapping is incorporated as an auxiliary objective within this multitask learning setup. The comprehensive methodology, depicted in Figure \ref{fig:conceptual_figure}, encompasses data pre-processing, feature extraction via VMamba \citep{liu2024vmamba} encoders, and multitask decoding to produce building localization masks, building damage levels, and floodwater maps. In the decoding stage, we introduce Feature-Fusion State Space (FFSS) to jointly decode multimodal features encoded by VMamba. Finally, building damage levels are integrated with georeferenced building footprints to generate detailed building-level damage maps.

\subsection{Preliminaries}
State-space models (SSMs), which originate from the Kalman filter \citep{kalman1960new}, transform a one-dimensional sequence, denoted as \(x(t) \in \mathbb{R}\), into an output response \(y(t) \in \mathbb{R}\), by means of a hidden state \(\textbf{h}(t) \in \mathbb{R}^N\). Such models are typically expressed as linear ordinary differential equations (ODEs) in the following form:
\begin{equation}\label{eq:ssm1}
\textbf{h}'(t) = \textbf{A}\textbf{h}(t) + \textbf{B}x(t),
\end{equation}
\raggedbottom
\begin{equation}\label{eq:ssm2}
y(t) = \textbf{C}\textbf{h}(t) + \textbf{D}x(t),
\end{equation}
\raggedbottom
where \(\textbf{h}'(t)\) is the derivative of the current state \(\textbf{h}(t)\), \(\textbf{A} \in \mathbb{R}^{N \times N}\), \(\textbf{B} \in \mathbb{R}^{N \times 1}\), \(\textbf{C} \in \mathbb{R}^{1 \times N}\), \(\textbf{D} \in \mathbb{R}\) are the learnable parameters. Eq. \ref{eq:ssm1} defines the state update function, where the matrix \(\textbf{A}\) governs the temporal evolution of the current state \(\textbf{h}(t)\), while the matrix \(\textbf{B}\) determines how the current input \(x(t)\) influences this state. Eq. \ref{eq:ssm2} represents the output function, where the matrix \(\textbf{C}\) maps the current state \(\textbf{h}(t)\) to the output, and the matrix \(\textbf{D}\) determines the direct influence of the current input \(x(t)\) on the output. \(\textbf{D}x(t)\) can be interpreted as a feed-through or skip connection from input to output and is often omitted in some model formulations.

For SSMs to be suitable for discrete token sequences and compatible with deep learning environments, they must be subjected to a discretization process that converts the continuous parameters into discrete counterparts. Typically, discretization techniques divide a continuous timeline into \(K\) distinct time intervals, ideally maintaining an equivalent integration area for every interval. A prominent method to accomplish this transformation is the zero-order hold (ZOH) method, which has been effectively applied in SSMs. The ZOH approach operates on the premise that the value of the function remains unchanged within each time interval, denoted as $ \Delta = [t_{k-1}, t_k] $. Following the application of ZOH discretization, Eq. \ref{eq:ssm1} and Eq. \ref{eq:ssm2} can be reformulated as
\begin{equation}\label{eq:ssm_discrete1}
\textbf{h}_k =  \overline{\textbf{A}}\textbf{h}_{k-1} + \overline{\textbf{B}}x_k,
\end{equation}
\raggedbottom
\begin{equation}\label{eq:ssm_discrete2}
y_k = \textbf{C}\textbf{h}_k,
\end{equation}
\raggedbottom
where
\begin{equation}\label{eq:ssm_discrete3}
\overline{\textbf{A}} = \text{exp}(\Delta\textbf{A}),
\end{equation}
\raggedbottom
\begin{equation}\label{eq:ssm_discrete4}
\overline{\textbf{B}} = (\Delta\textbf{A})^{-1}(\text{exp}(\Delta\textbf{A})-\textbf{I})\Delta\textbf{B}.
\end{equation}
\raggedbottom
Note that \(\textbf{D}x_k\) is omitted in this formulation.

In classical computer vision tasks, filters, typically in the form of convolution kernels, are utilized to derive aggregate features from input data. After undergoing discretization, SSMs can also be expressed as a convolutional operation. Given that SSMs inherently process sequential data, this convolutional representation employs a one-dimensional kernel. The parameters of this one-dimensional kernel are mathematically derived from the discretized SSM parameters. Specifically,
\begin{align}
& y_0  = \textbf{C}\textbf{h}_0 = \textbf{C}\overline{\textbf{B}}x_0,\\
& y_1  = \textbf{C}\textbf{h}_1 = \textbf{C}(  \overline{\textbf{A}}\textbf{h}_0 + \overline{\textbf{B}}x_1),\\
& y_k = \textbf{C}\textbf{h}_k = \textbf{C}(  \overline{\textbf{A}}\textbf{h}_{k-1} + \overline{\textbf{B}}x_k)
    = \textbf{C}\overline{\textbf{A}}^k \overline{\textbf{B}}x_0 + \textbf{C}\overline{\textbf{A}}^{k-1} \overline{\textbf{B}}x_1 + 
    \dots +
    \textbf{C}\overline{\textbf{A}}^1 \overline{\textbf{B}}x_{k-1} +
    \textbf{C} \overline{\textbf{B}}x_k.
\end{align}
\raggedbottom
Thus, the kernel \(\overline{\textbf{K}}\) can be formulated as
\begin{equation}\label{eq:ssm_kernel1}
\overline{\textbf{K}} = (\textbf{C} \overline{\textbf{B}}, \textbf{C} \overline{\textbf{A}}^1\overline{\textbf{B}}, \dots,  \textbf{C} \overline{\textbf{A}}^k\overline{\textbf{B}}, \dots,  \textbf{C} \overline{\textbf{A}}^{K-1}\overline{\textbf{B}}).
\end{equation}
\raggedbottom
The recurrent computation can now be converted to convolutional operation as
\begin{equation}\label{eq:ssm_kernel2}
\textbf{y} = \textbf{x}*\overline{\textbf{K}},
\end{equation}
\raggedbottom
where \(\textbf{x} = [x_0, x_1, \dots, x_{K-1}] \in \mathbb{R}^K\) and  \(\textbf{y} = [y_0, y_1, \dots, y_{K-1}] \in \mathbb{R}^K\) denote the input and output sequences.

A key limitation of prior SSMs is that only those exhibiting linear time invariance (LTI) can be efficiently computed as convolutions. Scenarios with variable spacing between inputs and outputs cannot be effectively modeled by static convolution kernels. However, the fixed dynamics of LTI models restrict their ability to perform content-based reasoning.  Mamba \citep{gu2023mamba} addresses these limitations in SSMs by introducing a selective scan mechanism.  This approach overcomes the constraints of LTI models by rendering SSM parameters input-dependent, thereby enabling selective information processing.  Additionally, Mamba employs a hardware-aware parallel algorithm to ensure computational efficiency.  Specifically, because Mamba's time-varying design precludes the use of efficient convolutions, it utilizes a parallel scan algorithm for its recurrent computations. This algorithm is optimized for graphics processing units (GPUs) through kernel fusion, which minimizes memory input/output operations, and employs recomputation during training to avoid storing intermediate states for backpropagation, thus managing the memory demands of its selective states.

\subsection{Network Architecture}
The proposed Flood-DamageSense model builds upon the VMamba \citep{liu2024vmamba} architecture and the ChangeMamba \citep{chen2024changemamba} framework. Figure \ref{fig:conceptual_figure} illustrates the proposed network architecture. The input data comprises pre- and post-event SAR intensity and InSAR coherence in vertical-vertical (VV) and vertical-horizontal (VH) polarizations, pre-event VHR optical satellite imagery, and a historical flood risk level map. Initially, input data are up-sampled to achieve a common resolution, matching that of the highest-resolution input, thereby ensuring consistency for subsequent operations. The encoding stage employs a semi-Siamese network structure featuring two distinct encoders based on the Visual State Space (VSS) backbone in VMamba: one designated as an image encoder and the other as a flood risk data encoder. Pre- and post-event SAR and InSAR images, along with pre-event VHR optical images, are processed by a single, weight-shared image encoder to facilitate the joint learning of spatial information. On the other hand, historical flood risk data are encoded separately by the flood risk data encoder, accounting for distinct feature patterns and lower information density of flood risk data compared to image data. Taking into account the differing channel counts of each sensor type, SAR/InSAR and optical image data are initially processed by individual convolutional embedding layers to ensure dimensional consistency prior to entering the image encoder.

\begin{figure}[ht]
    \centering
    \includegraphics[width=\textwidth]{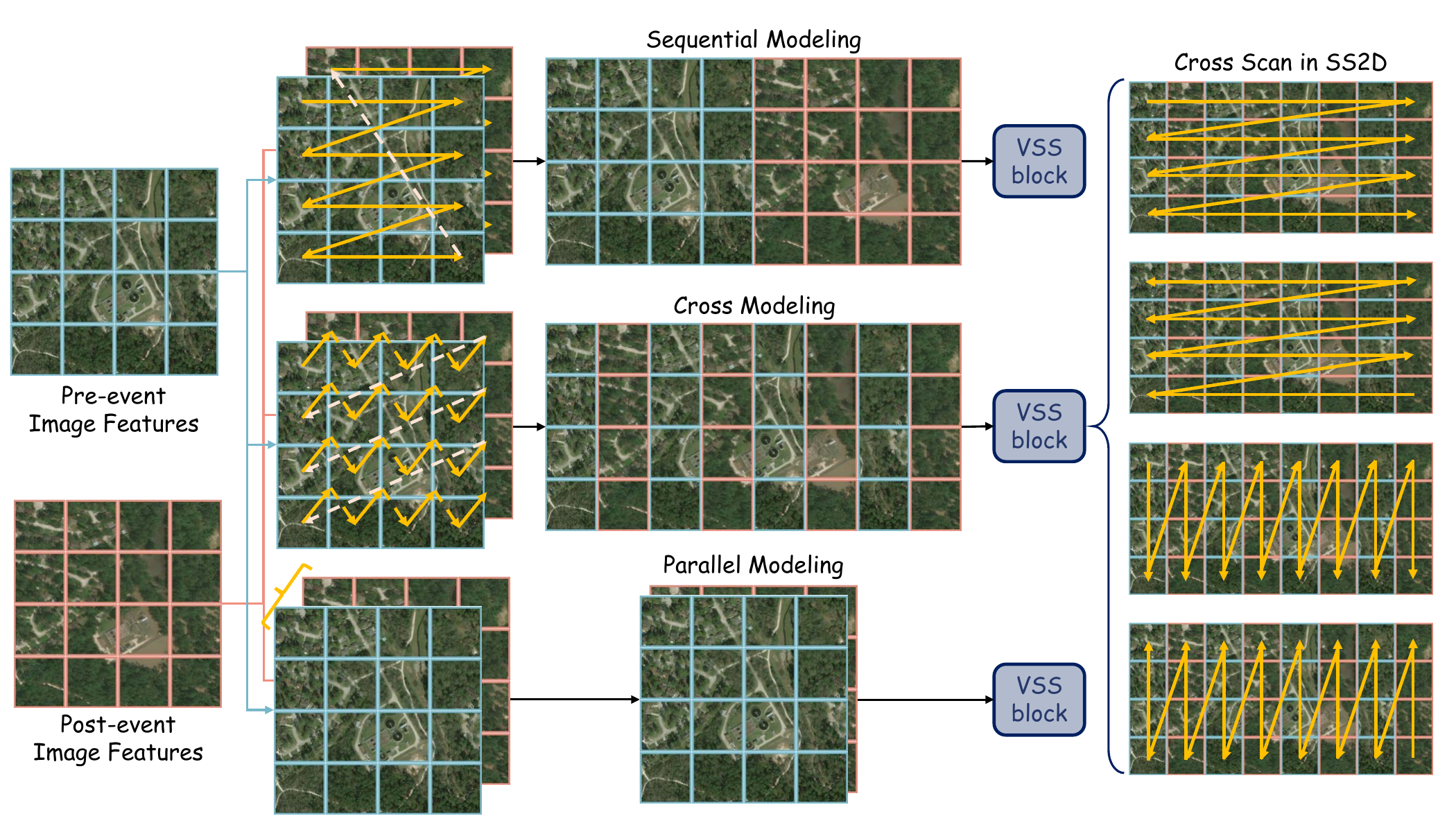}
    \caption{Illustration of Spatio-Temporal State Space. An STSS module contains three Visual State Space blocks. STSS processes pre- and post-event image features through a combination of sequential, cross, and parallel modeling mechanisms. The rearranged features by each modeling mechanism are then fed into a corresponding VSS block. In VSS, a cross-scan mechanism uses four-directional scanning paths to effectively model global contextual information from the spatio-temporal features. It should be noted that the depicted inputs represent encoded features instead of the original images. The optical satellite imagery shown, derived from xBD dataset \citep{Gupta_2019_CVPR_Workshops}, is used in the illustration only for enhancing conceptual understanding.}
    \label{fig:stss}
\end{figure}

In the decoding stage, this study introduces a Feature-Fusion State Space module to jointly decode the multimodal features. This FFSS module extends the concept of Spatio-Temporal State Space (STSS) from ChangeMamba, depicted in Figure \ref{fig:stss}, to accommodate multiple feature inputs. Three task-specific decoders then utilize different combinations of these processed features. Specifically,

\begin{itemize}
    \item The building damage decoder processes features from pre- and post-event SAR and InSAR imagery, pre-event VHR optical imagery, and historical flood risk levels as input to perceive the extent of building damage through spatio-temporal relationships and information unrelated to the event.
    \item The floodwater mapping decoder takes features from pre- and post-event SAR and InSAR imagery, functioning as a pure change detection decoder.
    \item The building localization decoder uses features from pre-event SAR and InSAR imagery and pre-event VHR optical imagery, operating as a semantic decoder with feature fusion capabilities.
\end{itemize}

While the building localization output is not a direct final product in this framework, it serves an auxiliary role in refining the association of damage assessments at the building level, particularly before the final aggregation using georeferenced building footprints. 

\subsubsection{Semi-Siamese Encoding}
\begin{figure}[ht]
    \centering
    \includegraphics[width=0.85\textwidth]{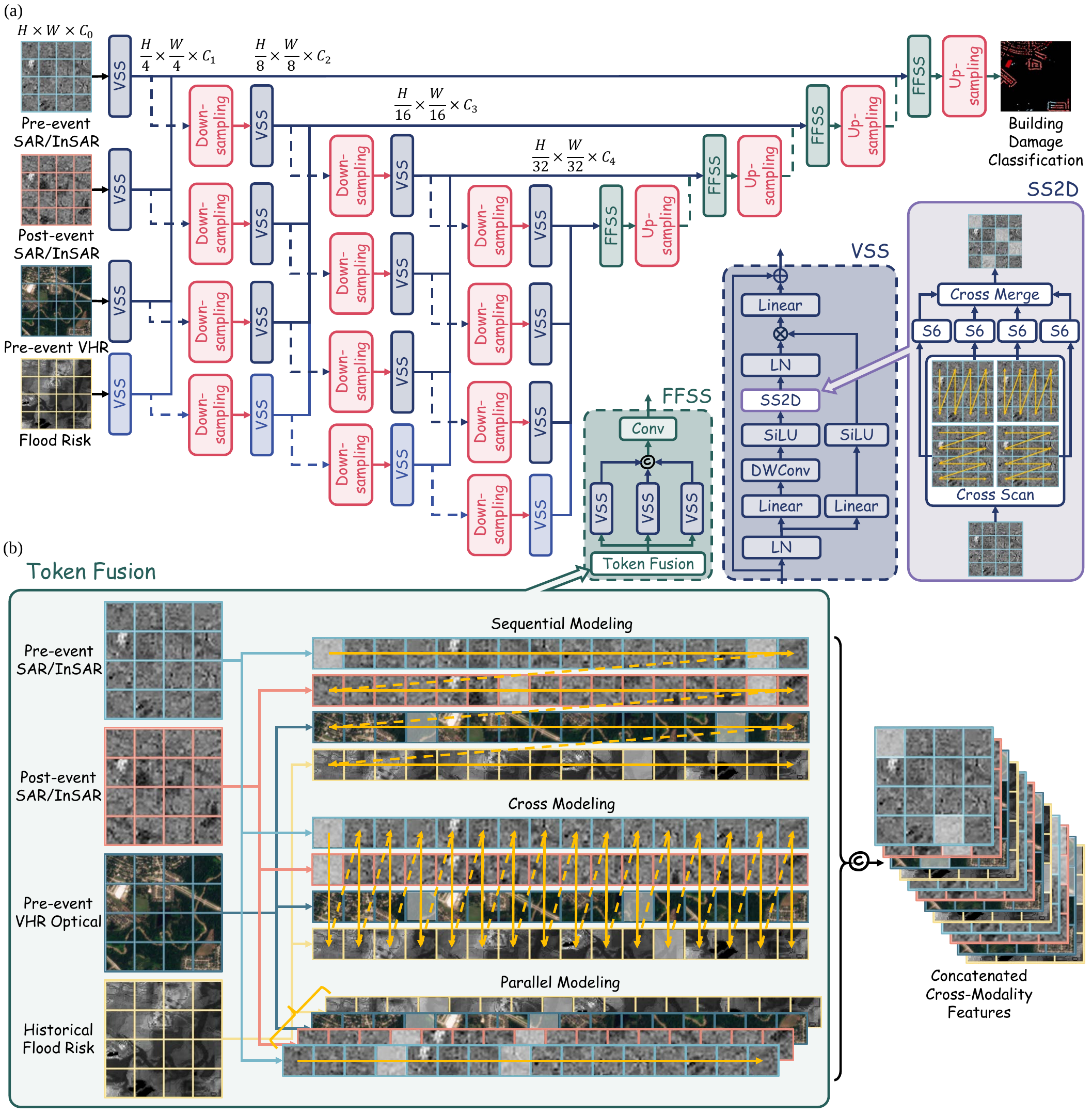}
    \caption{Encoder-decoder network and Feature-Fusion State Space details within the Flood-DamageSense model. (a) Detailed architecture of the encoder-decoder network. The encoder section features multiple parallel streams of Visual State Space blocks with progressively down-sampling to extract hierarchical features. Subsequently, the decoder employs blocks alongside up-sampling stages to reconstruct spatial details and produce the building damage classification output. Three expanded views detail the internal structures of an FFSS block, a VSS block, and the 2D-Selective-Scan  mechanism. While the diagram primarily illustrates the pathway for the building damage decoder, the depicted decoder architecture is adopted by all three task-specific decoders in the full model. (b) Conceptual diagram of token rearrangement mechanisms within the FFSS module. FFSS is designed to process VSS-encoded features derived from diverse input modalities. These features undergo sequential, cross, and parallel modeling approaches informed by spatio-temporal modeling techniques in ChangeMamba, which function as mechanisms for rearranging and fusing tokens from the various sources. The cross-modality features resulting from these three modeling mechanisms are subsequently concatenated. Specific SS2D operations are not detailed in this illustration, as their principles are elaborated in Figure \ref{fig:stss}. Furthermore, similar to the depiction in Figure \ref{fig:stss}, the inputs shown here represent VSS-encoded features rather than raw image data.}
    \label{fig:encoder_decoder_&_ffss}
\end{figure}

The Mamba architecture was initially developed for processing sequential data characterized by causal relationships. Consequently, its original one-dimensional scanning mechanism is not readily transferable to non-sequential data modalities, such as imagery, which typically exhibit spatial rather than strict causal dependencies. VMamba adapts the Mamba architecture for vision tasks by introducing the 2D Selective Scan (SS2D), which employs a four-way cross-scanning mechanism that transforms two-dimensional image patches into multiple flattened one-dimensional sequences by reordering tokens along four diverse paths. These token sequences are then independently processed by four different selective scan structured state space sequence (S6) blocks. The resulting features from each scan direction are subsequently merged, enabling the effective capture and reconstruction of comprehensive two-dimensional spatial context in the final feature map. The SS2D mechanism is detailed in the SS2D panel in Figure \ref{fig:encoder_decoder_&_ffss}(a).

Capitalizing on the VMamba architecture's effective adaptation of SSMs for vision tasks through the SS2D mechanism, the encoding stage of the proposed Flood-DamageSense model is specifically designed to process multimodal inputs and extract hierarchical features for subsequent damage assessment, as illustrated in the left half side of Figure \ref{fig:encoder_decoder_&_ffss}(a). This encoding stage utilizes a semi-Siamese network structure. Such a configuration incorporates a weight-sharing encoder and a feature-specific encoder, both founded on the Visual State Space backbone architecture derived from VMamba, detailed in the VSS panel in Figure \ref{fig:encoder_decoder_&_ffss}(a), where LN represents layer normalization, linear refers to a linear layer, while DWConv denotes a depth-wise convolution operation. The first component of this semi-Siamese network, designated as the weight-shared image encoder, processes a diverse range of image inputs, including pre- and post-event SAR and InSAR imagery, alongside pre-event VHR optical imagery. These varied inputs are initially processed by distinct embedding layers tailored to each specific sensor modality before being channeled into this single, weight-shared image encoder. This architectural choice is adopted because weight-sharing offers several distinct advantages: it ensures the extraction of consistent feature representations across different temporal states and sensor modalities, enhances parameter efficiency by reducing the total number of learnable parameters, thereby potentially mitigating overfitting, and directs the model to focus on discerning genuine spatio-temporal changes and inter-image relationships rather than variations arising from disparate processing pathways. The effectiveness of pairing distinct embedding layers designed to standardize dimensionality and to capture unique sensor-specific characteristics with a shared encoder aimed at learning joint representations has been demonstrated in multisensor geospatial models \citep{han2024bridging}. The second component, the flood risk data encoder, is specifically dedicated to processing historical flood risk data. This architectural separation accommodates the distinct feature patterns and typically lower information density inherent in flood risk data compared to the richer content of satellite imagery. The encoder network comprises four hierarchical stages designed to capture semantic information across multiple scales, ranging from global to local. The initial stage partitions the input image into patches. In contrast, each of the three subsequent stages commences by downsampling its input features. Following these initial operations, either patch partition or downsampling, every stage then employs a series of VSS blocks for comprehensive modeling of spatial contextual information, thereby producing hierarchical feature representations.

\subsubsection{Feature Fusion Decoding}
The decoding stage is crucial for synthesizing the rich, multimodal information generated by the VSS encoders. This complex integration is principally managed by the Feature-Fusion State Space module. The design of FFSS is inspired by the Spatio-Temporal State Space module in ChangeMamba, stemming from the conceptual similarity between fusing spatio-temporal information and fusing multimodal information, as both processes involve integrating features from disparate sources. Consequently, the FFSS module is engineered to concurrently process both spatio-temporal and multimodal information.

As illustrated in Figure \ref{fig:encoder_decoder_&_ffss}(b), the FFSS module receives VSS-encoded features derived from diverse modalities. Within this module, these multimodal features are subjected to an intricate integration process that employs three distinct modeling strategies to effectively reorder and fuse tokens from the various input streams. The token sequences, rearranged by these strategies, are then processed individually by three corresponding VSS blocks. The internal structure of an FFSS module, including these VSS components, is depicted in the FFSS panel in Figure \ref{fig:encoder_decoder_&_ffss}(a). As demonstrated in Figure \ref{fig:stss}(b), each of the three rearranged token sequences subsequently undergoes a four-way cross scan implemented by SS2D mechanism. The approach combining FFSS and SS2D can be conceptualized as a double-layer superposed scanning process, which enables the exploration of 12 types of cross-modality and spatio-temporal relationships. Consequently, this comprehensive scanning methodology, operating over multiple sequences and directions, significantly enhances the model's capacity to capture diverse contextual information. Subsequently, the outputs from these VSS blocks are subjected to a reverse rearrangement process and concatenated, yielding a unified and enriched representation of cross-modality features.

The decoder network mirrors the encoder's four-stage hierarchical structure, operating across corresponding scales, as depicted in the right side of Figure \ref{fig:encoder_decoder_&_ffss}(a). At each scale, VSS-encoded features are fed into a corresponding FFSS block. The token rearrangement mechanisms, illustrated in Figure \ref{fig:encoder_decoder_&_ffss}((b), are applied within each of the four decoding stages. The output from each stage's FFSS block is then integrated with the up-sampled output from the FFSS block of the preceding, deeper stage. The output from the final FFSS block constitutes the ultimate feature representation for the specific task, effectively integrating hierarchical information from all decoding stages.

The mechanisms described above pertain to the architecture of a single decoder pathway. The proposed Flood-DamageSense model incorporates three such task-specific decoders. While all share the same core FFSS-based architecture, they are trained with distinct combinations of encoded features tailored to their specific objectives. The building damage decoder utilizes all feature types to jointly learn spatio-temporal relationships and cross-modality representations, where the incorporation of event-irrelevant characteristics, such as historical flood risk, enhances the model's sensitivity to non-structural damage. The floodwater mapping decoder processes features from pre- and post-event SAR and InSAR imagery to conduct spatio-temporal change detection. The building localization decoder takes features from pre-event SAR/InSAR and VHR optical satellite imagery as inputs to perform semantic segmentation using multimodal information.

\subsection{Loss Function and Evaluation Metrics}
The Flood-DamageSense model is trained using a composite loss function, \(L^{final}\), which aggregates losses from three primary tasks. This multitask learning approach allows the model to learn shared representations while optimizing for each specific objective. For each task \(i \in \{BDA, FM, LOC\}\), representing building damage classification, floodwater mapping, and building localization, if valid ground truth labels are present, the loss \(L^i\) is computed as a weighted combination of cross-entropy (\(ce\)) loss and Lovasz-Softmax (\(lov\)) loss:
\begin{equation}\label{eq:task_loss}
L^i = \lambda^i_{ce} \cdot L^i_{ce} + \lambda^i_{lov} \cdot L^i_{lov},
\end{equation}
\raggedbottom
where \(L^i_{ce}\) is the cross-entropy loss for task \(i\), \(L^i_{lov}\)  is the Lovasz-Softmax loss for task \(i\), \(\lambda^i_{ce}\) and \(\lambda^i_{lov}\) are task-specific hyperparameters that balance the contribution of each loss component. The cross-entropy loss for task \(i\), \(L^i_{ce}\), measures the dissimilarity between the true class distribution and the estimated probability distribution from the model for each pixel. For a set of \(M^i_{valid}\) valid pixels and \(K_i\) classes in task \(i\), with \(y^i_{j, k}\) being the true label for pixel \(j\) and class \(k\) in one-hot encoding and \(p^i_{j, k}\) being the output softmax probability for pixel \(j\) and class \(k\), the cross-entropy loss is formulated as
\begin{equation}\label{eq:cross_entropy_loss}
L^i_{ce} = -\frac{1}{M^i_{valid}} \sum_{j=1}^{M^i_{valid}} \sum_{k=1}^{K_i} \left( y^i_{j,k} \cdot \log(p^i_{j,k}) \right).
\end{equation}
\raggedbottom
The Lovasz-Softmax loss \citep{berman2018lovasz} is a specialized loss function primarily designed for semantic segmentation tasks. Its development was motivated by the observation that traditional pixel-wise loss functions, such as cross-entropy, do not directly optimize for common segmentation evaluation metrics like the Intersection over Union (IoU), also known as the Jaccard index. While cross-entropy penalizes each misclassified pixel independently, the IoU metric evaluates the overlap between the predicted segmentation mask and the ground truth mask for each class, thus capturing the quality of the segmentation at a more structural or region-based level. A model trained solely with cross-entropy might achieve high pixel-wise accuracy but still produce segmentations with poor object boundaries or shapes, leading to suboptimal IoU scores. To bridge this gap, the Lovasz-Softmax loss serves as a direct, differentiable surrogate for the IoU metric in the context of multi-class classification.

To address class imbalance, particularly in the building damage classification and floodwater mapping tasks, class-specific weights are incorporated into both the cross-entropy and Lovasz-Softmax loss calculations. These weights are determined based on the inverse frequency of each class derived from the training dataset to give appropriate importance to minority classes. The weight \(w^i_k\) for task \(i\) and class \(k\) is formulated as
\begin{equation}\label{eq:class_weight}
w^i_k = \frac{N^i}{N^i_k},
\end{equation}
\raggedbottom
where \(N^i\) is the total number of instances for task \(i\) and \(N^i_k\) is the number of instances belonging to class \(k\) for task \(i\) in the training dataset. Class-specific weights are calculated at the building level, rather than at the more granular pixel level, thereby reducing the associated computational cost. Given Eq. \ref{eq:class_weight}, the weighted cross-entropy loss can be formulated as
\begin{equation}\label{eq:weighted_cross_entropy_loss}
L^i_{ce} = -\frac{1}{M^i_{valid}} \sum_{j=1}^{M^i_{valid}} \sum_{k=1}^{K_i} w^i_k\left( y^i_{j,k} \cdot \log(p^i_{j,k}) \right).
\end{equation}
\raggedbottom
The final composite loss \(L^{final}\) for training the model is the sum of the individual losses from the three tasks:
\begin{equation}\label{eq:final_loss}
L^{final} = L^{BDA} + L^{FM} + L^{LOC} = \lambda^{BDA}_{ce} \cdot L^{BDA}_{ce} + \lambda^{BDA}_{lov} \cdot L^{BDA}_{lov} + \lambda^{FM}_{ce} \cdot L^{FM}_{ce} + \lambda^{FM}_{lov} \cdot L^{FM}_{lov} + \lambda^{LOC}_{ce} \cdot L^{LOC}_{ce} + \lambda^{LOC}_{lov} \cdot L^{LOC}_{lov}.
\end{equation}
\raggedbottom
If no valid labels are present for task \(i\) for a given batch, the contribution of the loss of task \(i\), \(L^{i}\), to \(L^{final}\) is zero.

To assess the performance of the proposed model, particularly in the context of multi-class classification and segmentation tasks, a set of robust evaluation metrics is employed. The primary metric focused upon is the F1 score, calculated individually for each class, along with an aggregated measure using the harmonic mean of these per-class F1 scores. The F1 score for a specific class \(k\) is the harmonic mean of precision \(P_k\) and recall \(R_k\), providing a balance between these two often competing measures. It is particularly useful when dealing with imbalanced class distributions. Precision measures the accuracy of positive predictions, while recall measures the model's ability to identify all actual positives. Precision \(P_k\) and recall \(R_k\) for class \(k\) are defined as
\begin{equation}\label{eq:precision}
P_k = \frac{TP_k}{TP_k + FP_k},
\end{equation}
\raggedbottom
\begin{equation}\label{eq:recall}
R_k = \frac{TP_k}{TP_k + FN_k},
\end{equation}
\raggedbottom
where \(TP_k\), true positives for class \(k\), is the number of instances correctly classified as class \(k\), \(FP_k\), false positives for class \(k\), is the number of instances incorrectly classified as class \(k\), and \(FN_k\), false negatives for class \(k\), is the number of instances belonging to class \(k\) but incorrectly classified as other classes. The F1 score for class \(k\), \(F1_k\), is then calculated as
\begin{equation}\label{eq:f1}
F1_k = 2 \cdot \frac{P_k \cdot R_k}{P_k + R_k}.
\end{equation}
\raggedbottom
Calculating the F1 score for each class allows for a granular understanding of the model's performance across the different categories it is tasked to identify. To obtain a single, comprehensive indicator that summarizes the model's overall performance across all classes, the harmonic mean of the individual F1 scores is utilized. The harmonic mean is preferred over a simple arithmetic mean because it is more sensitive to lower F1 scores; it penalizes models that perform very poorly on one or more classes, even if they perform well on others, thus providing a more conservative and often more realistic measure of overall efficacy, especially in the presence of varying per-class performance. The harmonic mean of F1 scores for task \(i\) is given by
\begin{equation}\label{eq:f1_hmean}
F1^i_{HMean} = \frac{K_i}{\sum^{Ki}_{k=1}\frac{1}{F1^i_k}}.
\end{equation}
\raggedbottom
Notably, for the purpose of calculating an aggregate performance metric using the harmonic mean of F1 scores, the no damage class was excluded in the context of building damage assessment, and analogously, the non-flooded class was excluded for the floodwater mapping task. This methodological choice ensures that the evaluation focuses more critically on the model's proficiency in detecting and distinguishing between the actual presence and severity of damage or flooding; these classes are typically of greater operational significance and often represent more challenging classification scenarios. By excluding these often predominant negative classes, the aggregate harmonic mean F1 score provides a more stringent and representative measure of the model's performance on the target classes of interest, mitigating potential inflation of the metric due to high accuracy on easily identifiable no damage or non-flooded instances.

In addition to the standard F1 score, two modified F1 metrics, the adjacent F1 score and the upper adjacent F1 score, are also introduced to evaluate performance in building damage assessment, a task characterized by ordinal class relationships. These metrics are particularly suited for tasks with ordinal relationships as they offer a more nuanced understanding by considering predictions close to the true class as partially correct. Specifically, the adjacent F1 score for a given class \(k\), denoted as \(F1_{adj,k}\), relaxes the criterion for a correct prediction by assigning credit if the model's output is the true class \(k\) or its immediate neighbors (\(k-1\) or \(k+1\)). This approach is especially pertinent for ordinal scales where misclassifications between adjacent categories are often considered less critical than errors involving more distant classes. To calculate the adjacent F1 score \(F1_{adj,k}\), the precision and recall are redefined. Adjacent precision \(P_{adj,k}\) quantifies, for all instances predicted as class \(k\), the proportion that are correctly identified according to the adjacent criteria, which can be formulated as
\begin{equation}\label{eq:adj_precision}
P_{adj,k} = \frac{N(GT \in \{k-1, k, k+1\} \wedge Pred = k)}{N(Pred = k)}.
\end{equation}
\raggedbottom
Adjacent recall \(R_{adj,k}\) measures, for all instances where the ground truth is class \(k\), the proportion that are correctly identified by the model under the adjacent criteria, which can be formulated as
\begin{equation}\label{eq:adj_recall}
R_{adj,k} = \frac{N(GT = k \wedge Pred \in \{k-1, k, k+1\})}{N(GT = k)}.
\end{equation}
\raggedbottom
\(F1_{adj,k}\) is then computed using the standard F1 formula. The upper adjacent F1 score \(F1_{upper\_adj,k}\) is a stricter variation. It considers a prediction correct if it is the true class \(k\) or the immediately higher adjacent class \(k+1\), which means more severe in the context of damage. This metric can be useful in scenarios where avoiding underestimation of severity is more critical than slight overestimation. For this metric, upper adjacent precision \(P_{upper\_adj,k}\) quantifies, for all instances predicted as class \(k\), the proportion where the ground truth is either class \(k\) or \(k-1\), which can be formulated as
\begin{equation}\label{eq:upper_adj_precision}
P_{upper\_adj,k} = \frac{N(GT \in \{k-1, k\} \wedge Pred = k)}{N(Pred = k)}.
\end{equation}
\raggedbottom
Upper adjacent recall \(R_{adj,k}\) measures, for all instances where the ground truth is class \(k\), the proportion correctly recalled by the model as class \(k\) or the immediately higher class \(k+1\), which can be formulated as
\begin{equation}\label{eq:upper_adj_recall}
R_{upper\_adj,k} = \frac{N(GT = k \wedge Pred \in \{k, k+1\})}{N(GT = k)}.
\end{equation}
\raggedbottom
These ordinal-aware F1 scores offer deeper insights into model behavior beyond strict per-class F1 score, particularly for tasks like damage assessment, where class proximity has a meaningful interpretation.

\subsection{Building Damage Map Generation}
After Flood-DamageSense produces pixel-wise damage scores, those rasters are translated into a vector product by overlaying them on a georeferenced building-footprint layer that has been re-projected to the same coordinate reference system. For every footprint polygon, the workflow extracts the intersecting damage-class pixels and assigns the structure a single label equal to the median of those values—a statistic that is both robust to stray misclassifications and faithful to the ordinal nature of the damage scale. Alternative summaries (e.g., mean, mode, or maximum) can be selected through a simple configuration option when a different balance between under- and over-estimation is desired. During aggregation, the algorithm also records pixel-coverage counts and within-footprint label variance, flagging buildings that have inadequate coverage or high internal disagreement so that analysts can quickly review potential alignment errors or canopy occlusion. The enriched footprint layer, now carrying a definitive damage class together with confidence attributes, is exported as an open-standard vector file (GeoPackage or GeoJSON) with embedded styling rules, making it ready for immediate use in common GIS environments or for web-tile streaming. By deferring aggregation until this final step, the pipeline preserves the full semantic richness of the satellite imagery throughout processing while delivering an inspection-ready, building-level damage map suitable for rapid triage, claims prioritization, and recovery planning.

\section{Experiments and Results}
\label{sec:experiments and results}
This section provides a comprehensive evaluation of the proposed Flood-DamageSense model. The effectiveness in building-level flood damage assessment was investigated using a case study focused on the 2017 Hurricane Harvey in Harris County, Texas. The model's performance was rigorously evaluated using a multimodal dataset comprising pre- and post-event SAR and InSAR imagery, pre-event VHR optical imagery, and historical flood risk data, with ground truth established from flood insurance claim data. The evaluation framework addresses several key dimensions: quantitative performance measured by per-class F1 scores, adjacent F1 scores, upper adjacent F1 scores, and their harmonic mean across damage classes; comparative analysis against relevant baseline models; and ablation studies designed to ascertain the specific contributions of Flood-DamageSense's architectural components and input modalities. The subsequent subsections will systematically present the study area and datasets, benchmark against baseline models, detail the experimental configurations, report on quantitative and qualitative findings, and discuss the outcomes of the ablation studies.

\subsection{Study Area and Datasets}
\subsubsection{Study Area}
The validation of the Flood-DamageSense model was conducted using a case study centered on Harris County, Texas, during the 2017 Hurricane Harvey, a catastrophic and exceptionally significant weather event, characterized by unprecedented rainfall that led to extensive and devastating flooding, causing widespread and varied degrees of building damage throughout the region. The selection of this major disaster as the study area offers a robust and challenging real-world scenario for evaluating the capabilities of advanced damage assessment methodologies. Critically, the availability of comprehensive ground truth damage data and pertinent geospatial datasets for this specific event and location was a decisive factor in the selection for this research. The multimodal datasets subsequently employed for model training and evaluation are detailed below.

\subsubsection{SAR and InSAR Imagery}
The primary SAR and InSAR data for this study are sourced from the UrbanSARFloods dataset \citep{Zhao_2024_CVPR}. UrbanSARFloods is a benchmark dataset derived from Sentinel-1 Single Look Complex (SLC) imagery, which was specifically designed to advance large-scale floodwater mapping capabilities, with a particular emphasis on addressing the challenges of urban flood detection alongside open-area flood mapping. The dataset furnishes pre-processed Sentinel-1 data products with a spatial resolution of 20 meters, including pre- and post-event SAR intensity and InSAR coherence, available in both vertical-vertical  and vertical-horizontal polarizations, across 18 global flood events. All data products were partitioned into image chips, each with dimensions of 512×512 pixels. UrbanSARFloods provides corresponding labels for floodwater mapping, categorizing areas into non-flooded areas, flooded open areas, and flooded urban areas, designated with pixel values of 0, 1, or 2, respectively. These labels were generated through a combination of semi-automatic methods applied across all image chips for broad coverage and detailed manual annotations for selected subsets using high-resolution optical data to ensure accuracy. The floodwater labels were adopted as the ground truth labels for our auxiliary floodwater mapping task. The subset of UrbanSARFloods data corresponding to Hurricane Harvey is utilized in this study due to the availability of ground truth damage data. This subset consists of 1,296 images. Given that the UrbanSARFloods dataset was initially designed for broader floodwater mapping, each original 512×512 pixel image covers a relatively large geographic region. To adapt these images for fine-grained, building-level analysis, they were partitioned into a grid of 8×8 patches, resulting in sub-images of 64×64 pixels each. This pre-processing step generated a total of 82,944 image patches for model input. The selection of  the UrbanSARFloods dataset was primarily driven by its comprehensive offering of both SAR intensity and InSAR coherence data,  the latter being particularly crucial for identifying floodwater in complex urban built-up environments where intensity-based detection can be ambiguous.

\subsubsection{VHR Optical Imagery}
Pre-event very-high-resolution optical satellite imagery, with a spatial resolution of 0.5 meters, was obtained from \citet{Apollo_Mapping}. This VHR imagery, however, was available only for a limited portion of the study area. In instances where VHR optical data were unavailable for a given location, an empty image matrix populated with the value 255 was supplied as input to the model. The value 255 is consistently used across all data modalities in this study to represent missing or no-data pixels.

\subsubsection{Historical Flood Risk Level Map}
To incorporate an event-independent measure of baseline flood susceptibility, this study utilizes a historical flood risk level map developed by \citet{mobley2021quantification}, which provides a flood hazard layer for southeast Texas using a machine learning approach. The methodology employed a random forest classification model trained on an extensive, long-term record of National Flood Insurance Program flood claims in conjunction with a comprehensive set of high-resolution geospatial data, encompassing topographic variables, such as elevation, distance to coast and streams, and height above nearest drainage, hydrologic parameters, such as saturated hydraulic conductivity and Manning's roughness, and land cover characteristics, including imperviousness. The resulting data product provides a flood risk map at a resolution of 30 meters. The inclusion of this historical risk assessment is intended to enhance the Flood-DamageSense model's sensitivity, particularly for detecting damages that may not involve significant or overt structural changes.

\subsubsection{Ground Truth Data}
The ground truth labels for building damage assessment in this study are based on property-damage-extent (PDE) data from \citet{LIU2024FloodDamageCast}. The derivation of this PDE indicator involves several key data processing steps, including an initial focus on building structural damage, excluding damage to contents, and the application of feature capping and min-max normalization techniques to NFIP and IA program claim values. This normalization ensures consistency and comparability across these two diverse datasets before merging them. This processed PDE serves as the primary target variable for training and evaluating the Flood-DamageSense model. Utilizing PDE as ground truth addresses the challenges associated with the difficulty and potential inaccuracies of manual flood damage annotation, especially for large-scale events, by providing a more quantitative and systematically derived measure of damage. Noted that the PDE dataset is derived from actual claim records, meaning properties without claims are not included within this processed damage extent data. The continuous PDE values were categorized into three levels of damage extent using K-means clustering, with the optimal number of clusters determined by the elbow method. For the purposes of the Flood-DamageSense model, a four-category damage classification scheme is adopted, utilizing the three PDE-derived damage levels to represent minor damage, moderate damage, and major damage, designated with pixels value of 1, 2, or 3, respectively. The fourth category, no damage, designated with a pixel value of 0, is then defined for properties that exist within the study area but are not represented in the claim-based PDE dataset. Finally, georeferenced building footprints for Harris County, obtained from the Microsoft USBuildingFootprints dataset \citep{USBuildingFootprints}, is utilized to generate the final large-scale building flood damage map and to provide the ground truth labels for the auxiliary building localization task.

\begin{figure}[ht]
    \centering
    \includegraphics[width=0.7\textwidth]{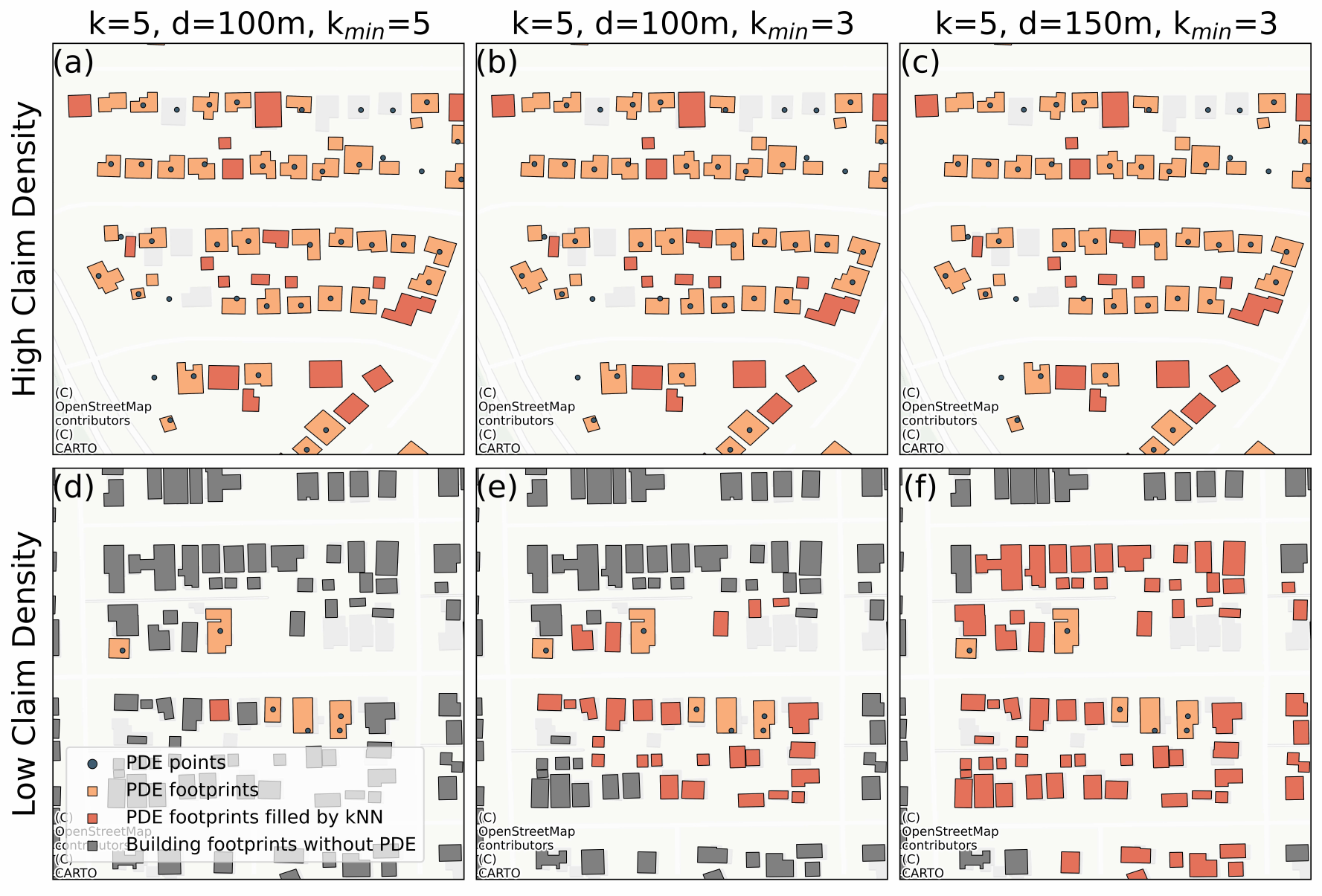}
    \caption{Visual comparison of kNN imputation outcomes for PDE across varying flood claim densities and kNN parameter settings. The top row (a–c) depicts results in a high claim density area, while the bottom row (d–f) corresponds to a low claim density area. The kNN configurations explored are: (a, d) \(k = 5\), \(d = 100\) m, and \(k_{min} = 5\); (b, e) \(k = 5\), \(d = 100\) m, and \(k_{min} = 3\); and (c, f) \(k = 5\), \(d = 150\) m, and \(k_{min} = 3\). Original PDE point data, building footprints with directly assigned PDE, building footprints with kNN-imputed PDE, and building footprints without PDE are presented.}
    \label{fig:pde_kNN}
\end{figure}

The initial PDE data, available as point data, underwent several pre-processing steps to be associated with individual building footprints. Firstly, PDE points were spatially joined to their corresponding building footprint polygons. For building footprints that did not directly contain any PDE points, an association was made with the closest PDE point, provided this distance did not exceed a 100-meter threshold, a measure implemented to minimize potential mismatches. To further account for buildings that did not have an initial flood claim but might have sustained damage, particularly if surrounded by properties with claims, a weighted average k-nearest-neighbors (kNN) imputation method was employed. The kNN approach was governed by three key parameters: \(k\), representing the number of nearest neighbors factored into the distance-weighted average; \(d\), defining the maximum distance limit for considering a neighbor; and \(k_{min}\), specifying the minimum number of neighboring properties with claims required to impute damage to an unclaimed property. The parameters \(d\) and \(k_{min}\) collectively dictate the criteria for identifying properties eligible for damage imputation. Three parameter configurations were subjected to experiment, as detailed in Figure \ref{fig:pde_kNN}. In areas characterized by high flood claim density, the three tested configurations produced comparable imputation results. However, in regions with low flood claim density, the two configurations utilizing a \(k_{min}\) value of 3 led to an excessive imputation of damage to properties. Based on these experimental observations, the parameter configuration of \(k = 5\), \(d = 100\) m, and \(k_{min} = 5\) was selected for the imputation of PDE values.

\subsection{Experimental Configurations}
\subsubsection{Data Preparation and Augmentation}
The dataset was split into training, validation, and test sets with ratios of 60\%, 20\%, and 20\%, respectively. For the proposed Flood-DamageSense network, as well as for models included in ablation studies, a consistent data preprocessing pipeline was adopted for multimodal inputs. Initially, all multimodal input data and their corresponding multitask labels were upsampled to a uniform resolution of 1280×1280 pixels to ensure consistency of resolution across the diverse data sources. During the training phase for these models, 512×512 pixel patches were then randomly cropped from these input data and labels. To enhance model robustness and generalization capabilities, several data augmentation techniques were applied to the training data, including random rotations, horizontal flips, and vertical flips. Input data were also normalized. For the baseline comparison models, the initial upsampling step was not necessary since these models utilize unimodal input. For these models, processing began with the original 512×512 pixel resolution of the input data, from which 256×256 pixel patches were randomly cropped for training.

\subsubsection{Training Parameters}
Using the AdamW optimizer, the Flood-DamageSense network, along with other networks used in baseline comparisons and ablation studies, was optimized. The initial learning rate was set to \(1\times10^{-4}\) and a weight decay of \(5\times10^{-3}\) was applied. For the composite loss function, the weights for the cross-entropy loss component for each task were set to 1.0; the weights for the Lovasz-Softmax loss component were set to 0.5 for the building localization and floodwater mapping tasks, following a similar strategic weighting approach as in ChangeMamba \citep{chen2024changemamba}. Recognizing building damage classification as the primary objective of Flood-DamageSense, the weight for its Lovasz-Softmax loss component was assigned a higher value of 0.75. All models were trained with an effective batch size of 16. Due to GPU memory limitations in certain experimental setups, this was implemented for some networks using an actual batch size of 2 complemented by 8 gradient accumulation steps or a batch size of 8 with 2 gradient accumulation steps. The training was conducted for a total of 200 epochs. During the testing and inference phase, the Flood-DamageSense model processed input data at its original, upsampled resolution of 1280×1280 pixels to generate predictions.

\subsection{Baseline Comparison}
To establish a robust foundation for the proposed Flood-DamageSense model, a comprehensive baseline comparison was conducted on the Hurricane Harvey subset of the UrbanSARFloods dataset. This evaluation considered representative models from three prominent deep learning paradigms: a convolutional-neural-network-based approach, Siamese-UNet \citep{SiameseUNet}; two Transformer-based approaches, BIT \citep{chen2022bit} and ChangeFormer \citep{bandara2022changeformer}; and a Mamba-based approach, ChangeMamba \citep{chen2024changemamba}. The input data were pre- and post-event SAR and InSAR imagery. To ensure suitability for this study's objectives, the decoder architectures of these baseline models were modified and adapted to the multitask learning framework. The quantitative performance of these adapted baseline models is presented in Table \ref{tab:baseline}.

\begin{table}[htbp]
\centering
\caption{Quantitative results of baseline models on the Hurricane Harvey subset of UrbanSARFloods dataset with additional building mask and damage labels. The harmonic mean of F1 scores excluded no damage class for building damage classification and non-flooded class for floodwater mapping.}
\resizebox{\textwidth}{!}{%
\begin{tabular}{@{}ccccccccccc@{}}
\toprule
 & \multirow{2}{*}{Batch} & \multirow{2}{*}{Building} & \multicolumn{5}{c}{Building Damage Assessment F1 Score} & \multicolumn{3}{c}{Floodwater Mapping F1 Score} \\
\cmidrule(lr){4-8} \cmidrule(lr){9-11}
Model & Size & Localization & No & Minor & Medium & Major & Harmonic & Open & Urban & Harmonic \\
& & F1 Score & Damage & Damage & Damage & Damage & Mean & Area & Area & Mean \\
\midrule
Siamese-Unet-ResNet-34 \citep{SiameseUNet} & 16 & 0.0031 & 0.9746 & 0 & 0 & 0.0035 & 0 & 0.3390 & 0.0002 & 0.0003 \\
Siamese-Unet-SE-ResNext-50 \citep{SiameseUNet} & 16 & 0.0048 & 0.9758 & 0 & 0 & 0.0058 & 0 & 0.3674 & 0 & 0 \\
Siamese-Unet-DPN-92 \citep{SiameseUNet} & 16 & 0.0692 & \textbf{0.9763} & 0 & 0 & 0 & 0 & 0.3861 & 0 & 0 \\
Siamese-Unet-SENet-154 \citep{SiameseUNet} & 16 & 0.4257 & 0.9711 & 0 & 0 & 0.1868 & 0 & 0.4177 & 0.0009 & 0.0017 \\
BIT-18 \citep{chen2022bit} & 16 & \textbackslash & 0.9692 & 0 & 0 & 0.1758 & 0 & 0.8145 & 0.0058 & 0.0116 \\
BIT-34 \citep{chen2022bit} & 16 & \textbackslash & 0.9711 & 0 & 0 & 0.2396 & 0 & 0.8332 & 0.2213 & 0.3498 \\
BIT-50 \citep{chen2022bit} & 16 & \textbackslash & 0.9708 & 0 & 0.0098 & 0.0978 & 0 & 0.5464 & 0.0057 & 0.0113 \\
ChangeFormerV1 \citep{bandara2022changeformer} & 16 & \textbackslash & 0.9713 & 0 & 0.0003 & 0.2581 & 0 & 0.8656 & 0.2161 & 0.3458 \\
ChangeFormerV2 \citep{bandara2022changeformer} & 16 & \textbackslash & 0.9719 & 0 & 0 & 0.2368 & 0 & 0.8606 & 0.1526 & 0.2592 \\
ChangeFormerV3 \citep{bandara2022changeformer} & 16 & \textbackslash & 0.9724 & 0 & 0 & 0.2525 & 0 & 0.8630 & 0.1326 & 0.2299 \\
ChangeFormerV4 \citep{bandara2022changeformer} & 16* & \textbackslash & 0.9735 & 0 & 0 & 0.2492 & 0 & \textbf{0.8816} & 0.3253 & 0.4753 \\
ChangeFormerV5 \citep{bandara2022changeformer} & 16* & \textbackslash & 0.9669 & 0 & 0 & 0.1062 & 0 & 0.8790 & 0.2974 & 0.4444 \\
ChangeFormerV6 \citep{bandara2022changeformer} & 16* & \textbackslash & 0.9722 & 0 & 0 & 0.2317 & 0 & 0.8780 & \textbf{0.3688} & \textbf{0.5194} \\
ChangeMamba \citep{chen2024changemamba} & 16 & \textbf{0.6319} & 0.9738 & \textbf{0.0012} & \textbf{0.0468} & \textbf{0.3132} & \textbf{0.0034} & 0.7998 & 0.2641 & 0.3971 \\
\bottomrule
\multicolumn{11}{l}{*Effective batch size is 16 and actual batch size is 8 with 2 accumulation steps due to limited available GPU memory.}

\end{tabular}
}
\label{tab:baseline}
\end{table}

Assessing building damage proved to be a demanding task for the majority of the baseline models, particularly for nuanced damage classes. Most Siamese-UNet, BIT, and ChangeFormer variants registered an F1 score of 0 for both minor damage and medium damage classes. While the F1 scores for no damage class were generally high across models, the F1 scores for major damage class varied. ChangeMamba (0.3132), BIT-34 (0.2396) and most ChangeFormer versions (e.g., V1: 0.2581, V3: 0.2525, V4: 0.2492) demonstrated capability in detecting major damage. Crucially, ChangeMamba was distinctive in achieving non-zero, albeit modest, F1 scores for minor damage (0.0012) and moderate damage (0.0468) classes. Consequently, its harmonic mean F1 score for damage assessment (0.0034) was more indicative of capability across multiple damage states compared to other models that often registered zero due to complete failures in intermediate classes. In the floodwater mapping task, several ChangeFormer variants,  ChangeMamba, and BIT-34 demonstrated competitive performance. ChangeFormerV4 achieved the highest F1 score for open area flooding (0.8816), and ChangeFormerV6 excelled in urban area flood mapping (0.3688), leading to the highest harmonic mean for this task (0.5194). The F1 scores for building localization reveal significant challenges for most baseline models. Most Siamese-UNet variants exhibited very low F1 scores, ranging from 0.003 to 0.4257. ChangeMamba demonstrated the best performance in this task, achieving a building localization F1 score of 0.6319. The building localization task was not performed in BIT and ChangeFormer since these models are pure change detection models without semantic decoders.

The comparative analysis indicates that while certain models, such as specific ChangeFormer variants, show strengths in auxiliary tasks like floodwater mapping, ChangeMamba achieves the best performance for building damage classification, which is our major task. In addition, ChangeMamba exhibited the most promising overall balance, particularly by demonstrating the capability in detecting all levels of building damage where other baselines largely failed. However, the generally low F1 scores for minor and medium damage classes across all baselines underscore the inherent lack of suitability of these tasks with existing architectures. This highlights a significant area for improvement and provides a key motivation for the development of the Flood-DamageSense model, which aims to enhance performance in these challenging aspects of building flood damage assessment. In addition, the results for building localization are consistent with the findings in extant literature that coarse spatial resolution of SAR imagery is insufficient for detailed building-level evaluation, underscoring the necessity of building footprint data to refine building-level damage results.

\subsection{Experimental Results}
This section presents the quantitative performance of the proposed Flood-DamageSense model on the Hurricane Harvey dataset. The evaluation focuses on its efficacy in building damage assessment at various levels of granularity, as well as its performance on the auxiliary tasks of building localization and floodwater mapping.

\subsubsection{Building Damage Assessment Performance}
The performance of the Flood-DamageSense model in building damage assessment, utilizing the configuration that incorporates all multimodal inputs, corresponding to configuration 4 in Table \ref{tab:ablation_study}, is detailed in Table \ref{tab:pixel_vs_building}. At the pixel level, the model achieved an F1 score of 0.9702 for the no damage class, indicating strong performance in identifying undamaged properties. For the actual damage classes, the F1 scores were 0.1109 for minor damage, 0.2100 for medium damage, and 0.3818 for major damage. The harmonic mean of F1 scores across minor, medium, and major damage classes at the pixel level was 0.1829. When aggregating the outputs to the building level, standard F1 scores improved for most damage classes, which were 0.9473 for no damage, 0.1255 for minor damage, 0.2487 for medium damage, and a notable 0.4610 for major damage. The harmonic mean of standard F1 scores for damage classes at the building level was 0.2629. This suggests that building-level aggregation, based on the median pixel value within a footprint, aids in refining the classifications. To account for the ordinal nature of damage assessment, building-level adjacent F1 scores, crediting predictions within \(\pm 1\) class of the true label, were computed. This metric yielded the F1 scores of 0.9840 for no damage, 0.9890 for minor damage, 0.4951 for medium damage, and 0.7545 for major damage. The harmonic mean for damage classes using this adjacent criterion was 0.7446. Although the F1 scores significantly increased with the adjacent consideration, double-sided adjacent consideration could raise concerns for the minor damage class, as classifying minor damage as no damage is generally undesirable in damage assessment. Furthermore, the upper adjacent consideration at the building level, crediting predictions that are the true class or the next higher damage class, resulted in the F1 scores of 0.9522 for no damage, 0.2387 for minor damage, 0.3923 for medium damage, and a notable 0.5999 for major damage. The harmonic mean for the damage classes with the upper adjacent consideration improved to 0.4230. The notable increase in this metric compared to the standard F1 score is a positive indicator, as a tendency to slightly overestimate damage is often preferable to underestimation in real-world disaster response and resource allocation, ensuring that potentially critical damage is not overlooked.

\begin{table}[htbp]
  \centering
  \caption{F1 scores at different levels for building damage assessment. The harmonic mean of F1 scores excluded no damage class.}
  \label{tab:pixel_vs_building}
  \resizebox{\textwidth}{!}{%
  \begin{tabular}{@{}lccccc@{}}
    \toprule
    Metric                 & No Damage & Minor Damage & Medium Damage & Major Damage & Harmonic Mean \\
    \midrule
    Pixel-level F1               & 0.9702    & 0.1109       & 0.2100        & 0.3818       & 0.1829        \\
    Building-level F1            & 0.9473    & 0.1255       & 0.2487        & 0.4610       & 0.2629        \\
    Building-level Adjacent F1   & 0.9840    & 0.9890       & 0.4951        & 0.7545       & 0.7446        \\
    Building-level Upper Adjacent F1 & 0.9522    & 0.2387       & 0.3923        & 0.5999       & 0.4230        \\
    \bottomrule
  \end{tabular}
  }
\end{table}

\begin{figure}[ht]
    \centering
    \includegraphics[width=\textwidth]{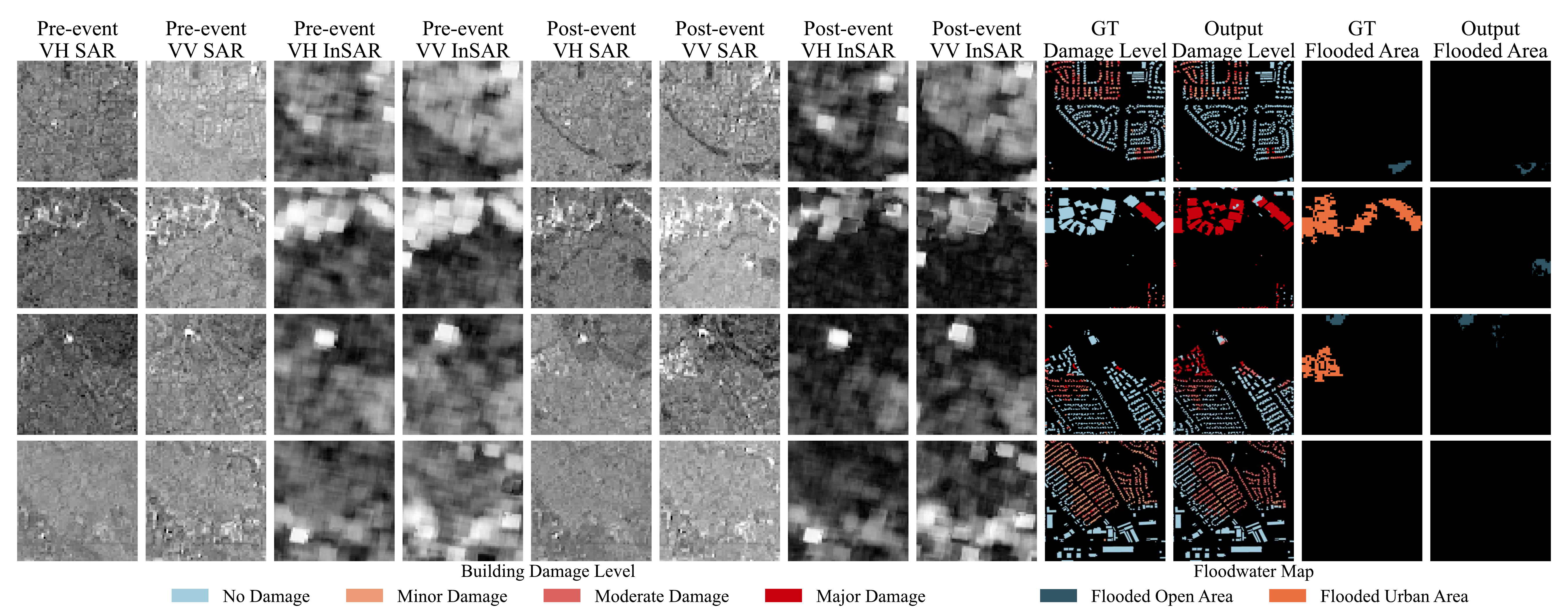}
    \caption{Examples of SAR and InSAR inputs and corresponding outputs for building damage assessment and floodwater mapping. The columns show pre- and post-event SAR and InSAR images (VH and VV polarizations), followed by ground truth (GT) and model outputs for building flood damage levels, categorized as no damage, minor, moderate, and major damage, and floodwater maps, distinguishing flooded open areas and flooded urban areas. Noted that for the visualization of building flood damage outputs, ground truth building masks, derived from the building footprint dataset, were utilized to delineate building locations, rather than masks generated by the model's auxiliary localization task in order to maintain consistency with the F1 score computation methodology for building damage assessment, which evaluates pixel-level classifications within known building extents, and because the final building damage map product of this framework relies on integrating damage outputs with existing, georeferenced building footprint data instead of concurrently produced building masks.}
    \label{fig:BDA_examples}
\end{figure}

\begin{figure}[ht]
    \centering
    \includegraphics[width=\textwidth]{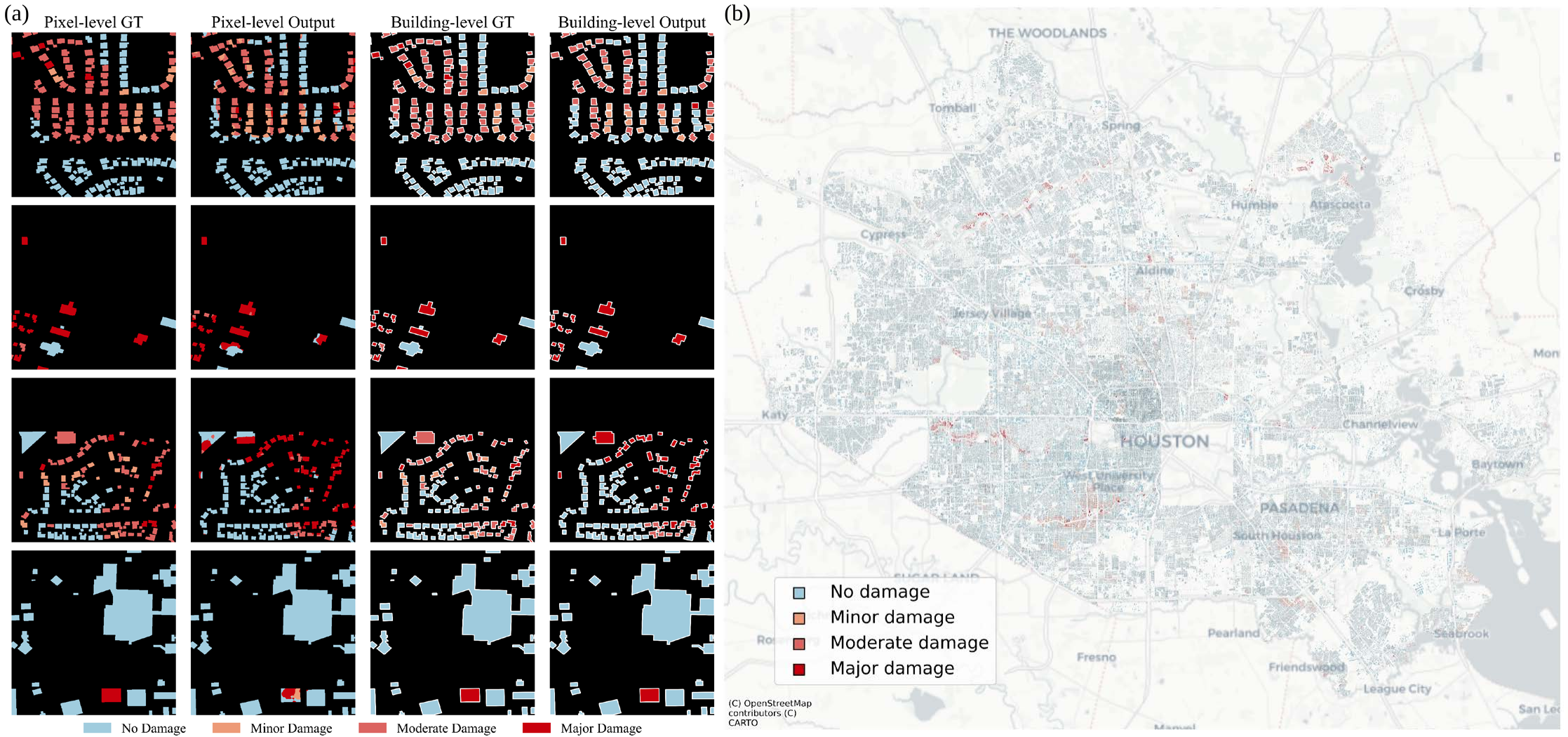}
    \caption{Demonstration of building-level flood damage outputs. (a) illustrates a detailed comparison of building flood damage outputs at pixel and building levels. (b) displays the comprehensive building-level flood damage map generated by Flood-DamageSense model for Harris County.}
    \label{fig:Building_Flood_Damage_Map}
\end{figure}

To complement these quantitative assessments, Figure \ref{fig:BDA_examples} presents qualitative examples of model-generated outputs for both building damage levels and floodwater mapping. The findings will be discussed alongside the results of floodwater mapping in Section \ref{sec:auxiliary_performance}. Figure \ref{fig:Building_Flood_Damage_Map} further demonstrates the proposed framework's capabilities in building damage assessment. Figure \ref{fig:Building_Flood_Damage_Map}(a) offers a granular comparison of building damage outputs at both the pixel level and the aggregated building level, which visually substantiates the advantage of the building-level processing strategy corresponding to Table \ref{tab:pixel_vs_building}. Figure \ref{fig:Building_Flood_Damage_Map}(b) presents the practical culmination of this analysis. The comprehensive building-level flood damage map generated by Flood-DamageSense for Harris County illustrates the framework's potential for producing large-scale, actionable damage assessments. For better visualization of map generation, both training and test data are presented in Figure \ref{fig:Building_Flood_Damage_Map}(b).

\subsubsection{Performance on Auxiliary Tasks}
\label{sec:auxiliary_performance}
The Flood-DamageSense model achieved a building localization F1 score of 0.3855. This performance is lower than the F1 score reported for the ChangeMamba model when operating with unimodal inputs, as detailed in Table \ref{tab:baseline}. A potential reason for this observed difference is the disparity in the resolution of the building mask labels used for evaluation in the respective experimental setups. Specifically, the building mask labels employed for evaluating Flood-DamageSense in this study possess a resolution 20 times higher than those used in the baseline comparison for ChangeMamba, even though the resolution of the SAR and InSAR input imagery remained consistent between these experiments. The building mask labels utilized in this research were derived from the building footprint dataset, which consists of vector data. A key characteristic of vector data is its scalability, allowing it to be rasterized to any desired resolution. Accordingly, for different experimental purposes within this study, the building footprints were rasterized at varying resolutions to align appropriately with the highest effective resolution of the input data in each specific configuration. Consequently, in the unimodal baseline comparison, the ChangeMamba model was tasked with delineating relatively coarse outlines, potentially encompassing groups of buildings. In contrast, the Flood-DamageSense model in this experiment was evaluated on its capability to achieve a more precise delineation of individual building boundaries, representing a considerably more challenging localization task due to the higher granularity of the ground truth. Beside, while pre-event VHR optical imagery was incorporated into Flood-DamageSense for building localization, its limited availability across the study area meant it could serve as auxiliary data rather than a consistently available primary input for building localization.

For floodwater mapping, the model yielded an F1 score of 0.8893 for flooded open area. However, the F1 score for flooded urban area was 0, resulting in a harmonic mean of 0 for the overall floodwater mapping task. This suggests that while proficient at detecting flooded open areas, the model struggled with urban floodwater detection. In addition to the quantitative metrics, Figure \ref{fig:BDA_examples} offers a qualitative illustration of the Flood-DamageSense model's performance. The figure displays side-by-side comparisons between ground truth data and the model's outputs for both building flood damage levels and floodwater delineation. Visual inspection of these outputs reveals a noteworthy observation. Although the floodwater mapping decoder failed to detect flooded urban areas, the building damage level outputs for these urban regions exhibited classifications that reflect the presence of inundation. Surprisingly, however, the corresponding ground truth PDE data for the buildings within these areas indicated no recorded economic loss. This observation underscores a crucial finding that the economic loss or damage extent captured by claim-based PDE data may not always directly correspond with the visually apparent presence of floodwater.

\subsection{Ablation studies}
To systematically evaluate the contributions of different input modalities and their specific application to the concurrent tasks within the multitask learning framework of Flood-DamageSense, a series of ablation studies were conducted. The performance of five distinct model configurations, as detailed in Table \ref{tab:ablation_study}, was analyzed. These configurations vary in their utilization of pre-event SAR/InSAR, post-event SAR/InSAR, pre-event VHR optical imagery, and historical flood risk data for the three concurrent tasks: building damage assessment (BDA), floodwater mapping (FM), and building localization (LOC).

\begin{table}[htbp]
\centering
\caption{Evaluation of the impact of multitask outputs and multimodal inputs on building localization, damage assessment, and floodwater mapping performance. The left half columns delineate which inputs were used in each task in each model. The middle column reports the batch size of input data in training process. The right half columns show the model performance for each task. The harmonic mean of F1 scores excluded no damage class for building damage classification and non-flooded class for floodwater mapping.}
\label{tab:ablation_study}
\resizebox{\textwidth}{!}{%
\begin{tabular}{@{}c ccc ccc ccc ccc | c | cccccc ccc@{}}
\toprule
\multirow{2}{*}{} & \multicolumn{3}{c}{Pre-event} & \multicolumn{3}{c}{Post-event} & \multicolumn{3}{c}{Pre-event} & \multicolumn{3}{c|}{} &   & Building & \multicolumn{5}{c}{ Building Damage Assessment (BDA)} & \multicolumn{3}{c}{Floodwater Mapping (FM)} \\
\multirow{2}{*}{No.} & \multicolumn{3}{c}{SAR/InSAR} & \multicolumn{3}{c}{SAR/InSAR} & \multicolumn{3}{c}{VHR} & \multicolumn{3}{c|}{Flood Risk} & Batch & Localization & \multicolumn{5}{c}{F1} & \multicolumn{3}{c}{F1} \\
\cmidrule(lr){2-4} \cmidrule(lr){5-7} \cmidrule(lr){8-10} \cmidrule(lr){11-13} \cmidrule(lr){16-20} \cmidrule(lr){21-23}
&   &   &   &   &   &   &   &   &   &   &   &   & Size & (LOC) & No & Minor & Medium & Major & Harmonic & Open & Urban & Harmonic \\
& LOC & BDA & FM & LOC & BDA & FM & LOC & BDA & FM & LOC & BDA & FM &   & F1  & Damage & Damage & Damage & Damage &  Mean & Area & Area & Mean \\
\midrule
0 & Y & Y &   &   & Y &   &   &   &   &   &   &   & 16  & 0.5750 & 0.9702 & 0      & 0.0121 & 0.3061 & 0      & \textbackslash & \textbackslash & \textbackslash \\
1 & Y & Y & Y &   & Y & Y &   &   &   &   &   &   & 16  & \textbf{0.6319} & 0.9738 & 0.0012 & 0.0468 & 0.2312 & 0.0034 & 0.7998 & \textbf{0.2641} & \textbf{0.3971} \\
2 & (Y)*  & Y & Y &   & Y & Y & Y &   &   &   & Y &   & 16* & 0.0395 & 0.9737 & 0.1135 & 0.1198 & 0.3196 & 0.1479 & 0.7772 & 0.1207 & 0.2090 \\
3 & Y & Y & Y &   & Y & Y &   &   &   &   & Y &   & 16* & 0.2759 & \textbf{0.9764} & 0.0909 & 0.1487 & 0.3302 & 0.1446 & 0.7969 & 0.0652 & 0.1840 \\
4 & Y & Y & Y &   & Y & Y & Y & Y &   &   & Y &   & 16* & 0.3855 & 0.9702 & \textbf{0.1109} & \textbf{0.2100} & \textbf{0.3818} & \textbf{0.1829} & \textbf{0.8893} & 0      & 0      \\
\bottomrule
\multicolumn{23}{l}{*Effective batch size is 16 and actual batch size is 2 with 8 accumulation steps due to limited available GPU memory.}\\
\multicolumn{23}{l}{*(Y) indicates pre-event SAR/InSAR data were used only when pre-event VHR optical imagery was not available.}
\end{tabular}
}
\end{table}

Model 0 is the original ChangeMamba model with unimodal inputs, which serves as a SAR-centric baseline. Model 0 utilizes pre- and post-event SAR/InSAR data for building damage assessment, and only pre-event SAR/InSAR data for building localization. This configuration does not perform the floodwater mapping task. Model 1 expands on Model 0 for multitask learning, using pre- and post-event SAR/InSAR data for both building damage assessment and floodwater mapping, while employing pre-event SAR/InSAR data for building localization. Model 1 is presented in Table \ref{tab:baseline}. Model 2 introduces multimodal inputs by incorporating historical flood risk data for building damage assessment. For building localization, it primarily uses pre-event VHR optical imagery, with pre-event SAR/InSAR data as a fallback where VHR imagery is unavailable. Model 3 is similar to Model 2 but relies solely on pre-event SAR/InSAR data for building localization, while still using SAR/InSAR with historical flood risk data for building damage assessment. Model 4 represents the full proposed Flood-DamageSense model, utilizing all input modalities for building damage assessment. For floodwater mapping, it uses pre- and post-event SAR/InSAR data. For building localization, it uses pre-event SAR/InSAR and pre-event VHR optical imagery. The primary task of building damage assessment benefits significantly from the integration of multiple data sources. Model 0 achieved a high F1 score for no damage and major damage but scored 0 for minor damage and very low for medium damage, resulting in a damage class harmonic mean F1 of 0. Model 1 showed similar performance for no damage and major damage but slightly higher for minor and medium damage, yielding the harmonic mean F1 of 0.0034. The introduction of the floodwater mapping task alongside building damage assessment enabled the model to achieve non-zero F1 scores across all damage categories. The introduction of historical flood risk data in Model 2 led to a substantial improvement in overall building damage assessment performance, with the harmonic mean F1 increasing to 0.1479. This was driven by notable gains in minor damage and medium damage F1 scores. Model 4, the full proposed model further incorporating pre-event VHR imagery for building damage assessment, achieved the highest harmonic mean F1 score of 0.1829 for damage classes. It demonstrated the strongest performance for all damage classes, although the F1 score for no damage was marginally lower than Model 3. The inputs for floodwater mapping remained consistent across all models. The performance of mapping flooded urban areas, however, decreased through the increase of building damage assessment performance. Model 4, despite having the highest F1 for flooded open areas, completely failed to detect flooded urban areas, resulting in a harmonic mean F1 of 0. The degradation in urban flood mapping performance as building damage assessment performance improved suggests a potential interplay or resource contention between the decoders. It is hypothesized that the building damage decoder in more complex configurations might learn features that overlap with the cues necessary for the floodwater mapping decoder to detect urban floods, a finding consistent with qualitative observations from Figure \ref{fig:BDA_examples}. Performance on the building localization task highlighted the importance of both input modality selection and multimodality fusion architecture. Model 2, which primarily utilized pre-event VHR imagery for localization with SAR/InSAR as a fallback where VHR was unavailable, experienced a drastic drop in F1 score to 0.0395. This suggests challenges in effectively utilizing the conditionally available VHR data or fusing it with SAR features in that specific setup. In this model configuration, the building localization decoder receives features that alternately originate from VHR optical imagery or, where VHR data are unavailable, from SAR imagery. Processing potentially disparate feature types through a common decoder not explicitly designed for adaptive multimodality fusion may present challenges in achieving optimal performance. Model 3, using pre-event SAR/InSAR data for localization, achieved an F1 score of 0.2759. Model 4, the full proposed model, which employed both pre-event SAR/InSAR and pre-event VHR imagery for localization using the specialized feature fusion mechanisms in the decoder, achieved an F1 score of 0.3855. An explanation for the comparatively lower F1 scores exhibited by Models 2-4, relative to Models 0 and 1, is provided in Section \ref{sec:auxiliary_performance}.

\section{Discussion}
\label{sec:discussion}
The comparative experiments highlight both the promise and the remaining hurdles of multimodal, multitask learning for rapid flood-damage intelligence. Ablation tests show that SAR/InSAR alone rarely differentiates minor and moderate losses; however, incorporating an inherent-risk prior doubles performance in those classes, and introducing pre-event VHR imagery lifts overall accuracy by sharpening medium- and high-damage predictions. The gains confirm that the all-weather reach of SAR must be complemented by contextual cues—historical exposure, land use, and fine-scale building texture—to capture the low-contrast signatures of inundation. Converting noisy pixel scores to a single, median damage label per footprint raises the harmonic-mean F1 again and achieves an upper-adjacent F1 of 0.60, ensuring that severely affected structures are seldom misranked—an outcome critical for triage and resource allocation. The auxiliary objectives expose subtler dynamics. Building localization is most accurate when a dedicated Feature-Fusion State Space module combines SAR and VHR streams; conditional fall-backs to SAR where VHR is missing degrade performance; and the task remains harder than in optical-only baselines because the evaluation mask is an order of magnitude more detailed. Floodwater mapping shows the opposite pattern: despite an unchanged floodwater mapping decoder architecture, as additional modalities and more complex decoders are introduced to improve damage grading, urban inundation detection deteriorates, even though open-area performance remains strong. Visual inspection suggests that the damage decoder monopolizes urban water cues, pointing to representational competition inside the shared encoder—a known risk of multitask networks that merits architectural or loss-balancing refinements. Qualitative examples also reveal a systematic gap between visible water and claim-based property-damage-extent labels: some urban blocks appear flooded yet register no economic loss, likely because of low insurance uptake or sub-threshold damage. While this mismatch can suppress F1 scores relative to purely visual ground truth, PDE remains the most scalable and practically relevant proxy for post-event recovery costs; its use therefore aligns the model with the priorities of emergency managers, albeit at the cost of a noisier training signal.

Three limitations warrant future work. First, claim dependency may under-represent damage to uninsured or under-insured buildings; supplementing PDE with crowd-sourced or governmental inspection data could mitigate bias. Second, sparse VHR coverage restricts the power of optical cues; lightweight self-supervised pre-training on global archives or diffusion-based up-sampling could increase effective coverage. Third, the failure to capture complex urban flood signatures indicates that SAR-specific physics priors or higher-resolution interferometric stacks are needed. Addressing these issues, and refining loss-sharing among decoders, should enable Flood-DamageSense to generalize across diverse hydrologic settings and deliver uniformly high performance across all subtasks. Overall, the study demonstrates that fusing complementary modalities through a Mamba backbone, supervising with economically meaningful labels, and aggregating predictions at the building footprint can produce actionable flood-damage maps within hours of image acquisition—marking a tangible advance toward operational, structure-scale disaster analytics.

\section{Concluding Remarks}
\label{sec:conclusion}
Flood-DamageSense advances the state of remote-sensing analytics in three fundamental ways. Methodologically, it is the first flood-specific, multi-level building-damage model that (i) exploits the linear-complexity Mamba sequence backbone to capture long-range spatial dependencies without the computational overhead of attention, (ii) fuses pre-/post-event SAR–InSAR, pre-event VHR optical imagery, and an inherent flood-risk prior within a semi-Siamese Visual State Space encoder and Feature-Fusion decoder, and (iii) learns three tightly coupled tasks—damage gradation, floodwater extent, and building localization—under a single multitask regime. This design overcomes the twin barriers that have hampered earlier work: cloud-induced data gaps and the subtle spectral signature of water damage that thwarts generic change-detection pipelines. Empirically, the Hurricane Harvey case study confirms that integrating a risk prior and SAR data produces large, statistically significant gains, particularly in the “minor” and “moderate” bins that drive most recovery-cost decisions but are routinely misclassified by xBD-era networks. Ablation results further reveal how each modality contributes—and occasionally competes—within the multitask framework, offering a blueprint for future architecture tuning.

From a practical perspective, the study delivers a full image-to-map workflow that translates pixel-wise probabilities into inspection-ready, geo-referenced vector layers using a median-aggregation and quality-flag scheme. This pipeline can ingest outputs from any pixel-level model, enabling agencies to plug in alternative classifiers while retaining the map-production backbone. Because the only post-event inputs are freely available SAR scenes, the system can operate within hours of a flood peak, providing site-level damage intelligence for triage routing, claims prioritization, and recovery-budget estimation—capabilities that go well beyond the binary, image-patch products found in prior flood studies.

Despite these advancements, this study acknowledges limitations such as the inherent nature of claim-based PDE ground truth, the restricted availability of VHR optical data, the challenges in robust urban flood mapping, and the need for broader testing for generalizability. Future research should aim to address these limitations by exploring enhanced multimodality fusion techniques, incorporating alternative or supplementary ground truth sources, and evaluating the model's performance across a wider array of flood events and diverse geographical settings. Further investigation into optimizing the interplay between multitask decoders is also warranted to ensure robust performance across all concurrent objectives. Flood-DamageSense represents a significant step towards more reliable and rapid building flood damage assessment, offering valuable support for post-disaster response and recovery efforts.

\section*{Data Availability}
UrbanSARFloods dataset is publicly available at \url{https://github.com/jie666-6/UrbanSARFloods}. The other data that has been used are confidential.

\section*{Code Availability}
The source code of the proposed model in this study can be found at \url{https://github.com/violayhho/Flood-DamageSense}.

\section*{Acknowledgments}
The authors would like to thank Dr. Samuel D. Brody (Texas A\&M University at Galveston) for providing invaluable historical flood risk data and our fellow Urban Resilience Lab member Dr. Chia-Fu Liu (Texas A\&M University) for providing property-damage-extent data for model development. The authors would also like to acknowledge funding support from the National Science Foundation under CRISP 2.0 Type 2,  grant 1832662, and the Texas A\&M University X-Grant 699. Any opinions, findings, conclusions, or recommendations expressed in this research are those of the authors and do not necessarily reflect the view of the funding agencies.

\bibliography{ref}

\end{document}